\documentclass[conference]{IEEEtran}
\IEEEoverridecommandlockouts
\usepackage{cite}
\usepackage{amsmath,amssymb,amsfonts}
\usepackage{algorithmic}
\usepackage{graphicx}
\usepackage{textcomp}
\usepackage{xcolor}
\def\BibTeX{{\rm B\kern-.05em{\sc i\kern-.025em b}\kern-.08em
    T\kern-.1667em\lower.7ex\hbox{E}\kern-.125emX}}

\newcommand{\bbC}{{\mathbb{C}}}

\title{Toward Data-Driven STAP Radar\\
}

\author{
\IEEEauthorblockN{Shyam Venkatasubramanian}
\IEEEauthorblockA{\textit{Duke University} \\
Durham, NC, USA
}
\and
\IEEEauthorblockN{Chayut Wongkamthong}
\IEEEauthorblockA{\textit{Duke University} \\
Durham, NC, USA 
}
\and 
\IEEEauthorblockN{Mohammadreza Soltani}
\IEEEauthorblockA{\textit{Duke University} \\
Durham, NC, USA
}
\and 
\IEEEauthorblockN{Bosung Kang}
\IEEEauthorblockA{\textit{University of Dayton}\\
Dayton, OH, USA
}
\and
\IEEEauthorblockN{Sandeep Gogineni}
\IEEEauthorblockA{\textit{ISL Inc} \\
San Diego, CA, USA
}
\and
\IEEEauthorblockN{Ali Pezeshki}
\IEEEauthorblockA{\textit{Colorado State University} \\
Fort Collins, CO, USA
}
\and 
\IEEEauthorblockN{Muralidhar Rangaswamy}
\IEEEauthorblockA{\textit{Air Force Research Laboratory}\\
Dayton, OH, USA
}
\and
\IEEEauthorblockN{Vahid Tarokh}
\IEEEauthorblockA{\textit{Duke University} \\
Durham, NC, USA
}
}

\begin{document}

\maketitle

\begin{abstract}
Using an amalgamation of techniques from classical radar, computer vision, and deep learning, we characterize our ongoing data-driven approach to space-time adaptive processing (STAP) radar. We generate a rich example dataset of received radar signals by randomly placing targets of variable strengths in a predetermined region using RFView, a site-specific radio frequency modeling and simulation tool developed by ISL Inc. For each data sample within this region, we generate heatmap tensors in range, azimuth, and elevation of the output power of a minimum variance distortionless response (MVDR) beamformer, which can be replaced with a desired test statistic. These heatmap tensors can be thought of as stacked images, and in an airborne scenario, the moving radar creates a sequence of these time-indexed image stacks, resembling a video. Our goal is to use these images and videos to detect targets and estimate their locations, a procedure reminiscent of computer vision algorithms for object detection---namely, the Faster Region Based Convolutional Neural Network (Faster R-CNN). The Faster R-CNN consists of a proposal generating network for determining regions of interest (ROI), a regression network for positioning anchor boxes around targets, and an object classification algorithm; it is developed and optimized for natural images. Our ongoing research will develop analogous tools for heatmap images of radar data. In this regard, we will generate a large, representative adaptive radar signal processing database for training and testing, analogous in spirit to the COCO dataset for natural images. Subsequently, we will build upon, adapt, and optimize the existing Faster R-CNN framework, and develop tools to detect and localize targets in the heatmap tensors discussed previously. As a preliminary example, we present a regression network in this paper for estimating target locations to demonstrate the feasibility of and significant improvements provided by our data-driven approach.
\end{abstract}


\section{Introduction} \label{Sec1}

The performance limits of STAP radar have been widely studied in the past (see \cite{murali_stap02,melvin_stap96,guerci_stap00}). Broadly speaking, three issues degrade clutter suppression, target detection, and target localization. (1) Interference due to clutter is unknown a priori and must be estimated from available data. Traditionally, the clutter covariance matrix is estimated from a limited number of returns from a radar dwell, whereby the accuracy of this estimation governs the performance of STAP \cite{guerci2003space}. When the dimensionality of the STAP weight vector grows large, so does the amount of data required for obtaining a good estimate of the clutter covariance matrix. A fundamental assumption used in covariance estimation methods is that the clutter scattering function is a wide-sense stationary process. This assumption only holds over relatively short dwell times, which limits the amount of data that can be used to build an estimate of the clutter covariance matrix. Consequently, there is always a mismatch between the true clutter statistics and the estimated values. As the dimensionality of the STAP weight vector increases to meet the ever increasing demand for resolution, this mismatch problem becomes more severe (see \cite{murali_stap02,melvin_stap96}). (2) When a target lies close to the clutter, there is always a considerable degree of overlap between the azimuth-Doppler subspace of the target response and that of the clutter response. The subspace leakage from the clutter results in a significant increase in the false alarm rate. Projecting the clutter response out (through either orthogonal or oblique projections) inadvertently results in removing some of the target response, leading to a reduction in the probability of correct detection (see  \cite{murali_stap02,melvin_stap96}). (3) Fully adaptive STAP radar, where a separate adaptive weight is applied to all elements and pulses, requires solving a system of linear equations of size $M \times N$. For typical radar systems, the product $M \times N$ may vary from several hundreds to tens of thousands. The sheer computational power needed to solve these large systems of equations within short time intervals to achieve real-time operation, especially for high-sampling rate radars (see \cite{guerci_stap00}), may be prohibitive.

The recent emergence of site-specific high-fidelity radio frequency modeling and simulation tools (such as RFView \cite{rfview}) has made it possible to take a data-driven approach to STAP radar. This has motivated our ongoing effort to address the aforementioned challenges. In our approach, received radar data is generated using RFView for random target locations and strengths within a constrained area. In this constrained area, we produce heatmap tensors of the desired test statistic or the output power of a selected beamformer in range, azimuth, and elevation angle (or other coordinates). Our goal is to use these heatmap tensors to detect targets and estimate their locations. Our approach is inspired by computer vision algorithms for object detection such as the Faster R-CNN \cite{ren2015faster}. The Faster R-CNN consists of a proposal generating network for determining regions of interest (ROI), a regression network for positioning anchor boxes around targets, and an object classification algorithm. It was mainly developed and optimized for natural images. Our ongoing research will develop analogous tools for the aforementioned heatmap images of radar data. In doing so, we will be generating a large, representative radar database for training and testing, which is in spirit analogous to the COCO dataset for natural images \cite{lin2014microsoft}. By building upon, adapting, and optimizing the existing components of Faster R-CNN, we will develop tools to detect and localize targets in the heatmap tensors mentioned above. As a preliminary example, we present a regression network in this paper for estimating target locations within a constrained area to demonstrate the feasibility of and significant improvements provided by our data-driven approach. In this preliminary example, the airborne radar is stationary with respect to the radar scene, and we use heatmaps of the output power of an MVDR beamformer \cite{vanveen_el88}.

The outline of the paper is as follows. In Section \ref{Sec2}, we briefly review RFView, a site-specific high-fidelity radio frequency modeling and simulation tool that is used to generate the rich data set required for our research. In Section \ref{Sec3}, we review the MVDR beamforming procedure for generating heatmap images for our training and test datasets. In Section \ref{Sec4}, we describe our example scenario in RFView which is used to generate the results in Section \ref{Sec6}. In Section \ref{Sec5}, we briefly describe our ongoing research and its relationship to existing computer vision object detection algorithms. In Section \ref{Sec6}, we present a regression convolutional neural network for target localization, and we provide numerical results demonstrating the performance and significant improvements of our proposed regression CNN scheme over the classical method. Finally, in Section \ref{Sec7}, we conclude with general takeaways.

\section{RFView Modeling and Simulation Tool and STAP Radar Database} \label{Sec2}

We first describe the RFView radio frequency (RF) modeling and simulation environment, which we will be using to generate a large, representative STAP radar database.

\subsection{RFView Modeling and Simulation Tool}

RFView is a site-specific modeling and simulation environment developed by ISL Inc. With this simulation platform, one can generate accurate radar data to be used for various signal processing algorithm applications. The software is built on a Splatter, Clutter, and Target Signal (SCATS) phenomenology engine that has successfully supported numerous advanced development projects for various organizations, including the U.S. Air Force and Navy, since 1989. This SCATS model is one of the first analysis tools to accurately characterize complex RF environments and perform several signal processing tasks, including system analysis and high-fidelity data generation. Examples of tools incorporated into the RFView model are target returns, direct path signal, ground scattered signal, and interference effects. To specify simulation scenarios and parameters in RFView, one can use either a MATLAB package or access a cloud-based web interface, run on a remote, high-speed computer cluster. A world-wide database of terrain and land cover data is also provided with the software package. 

The RFView platform provides many radar simulation capabilities, including high fidelity electromagnetic propagation modeling, multi-channel and MIMO radar simulation, multi-static clutter modeling, high-fidelity RF system modeling, channel mismatch modeling, and post-processing pipelines to analyze generated data. There are three main categories of parameters that users can adjust. The first is the platform and target structure, which includes the locations and the trajectories of the radar platforms as well as multiple target characteristics such as speed and radar cross-section (RCS). The second functionality is the task scheduler, which controls specific radar parameters such as the range swath, the pulse repetition frequency, the bandwidth, and the number of pulses. One can also control other receiver-specific parameters involving non-cooperative emitters. The third category is the antenna structure, which defines parameters pertaining to multi-channel planar arrays for both receivers and transmitters. Apart from these three categories, RFView also provides advanced simulation options, such as intrinsic clutter motion (ICM) modeling and parallel execution in `cluster mode' to facilitate reduced simulation runtimes.

\subsection{STAP Radar Database}

An integral part of our data-driven approach is the development of algorithms inspired by computer vision. Because of decades of intensive work, there now exist several publicly available datasets of natural images such as CIFAR-10/100~\cite{krizhevsky2009learning}, ImageNet~\cite{ILSVRC15}, and COCO~\cite{lin2014microsoft} for the training and testing of various computer vision algorithms. However, it appears that there is a lack of available STAP airborne radar datasets for data-driven research. In this effort, we plan to use RFView to fill this vacuum by generating a large, representative STAP radar database. To build this database, we will:
\begin{list}{}{\leftmargin=1em \itemindent=0em}
    \item \textbf{Step I}: Identify a set of scenarios, environments (over varied elevations and topographies), antenna array configurations, platform-to-scene distances, etc. that are representative of real-world situations encountered in STAP radar.
    \item \textbf{Step II}: Use RFView to generate radar data for randomly placed target(s) with prototypical radar cross-sections within a constrained area. The placement of these targets will be governed by the parameterized set given in Step I.
    \item \textbf{Step III}: Organize the results into an extensive database for data-driven STAP radar research. This database will consist of datasets containing image tensors and videos corresponding to beamformer output power and test statistic heatmaps of the simulated radar data.
\end{list}
To efficiently compile this database, we will be parallelizing the data generation process using RFView's built-in message passing interface (MPI). Among such features, RFView enables users to parallelize simulations over transmit pulse trains and/or clutter patches, an additional capability we will use to generate heatmaps of the output power of select beamformers and desired test statistics for our extensive database. The above steps will be thoroughly expanded upon in our future journal work detailing this STAP radar database. 


  
\section{Heatmap Images from MVDR Beamforming} \label{Sec3}

\begin{figure}[hbt!]
    \centering
    \includegraphics[width=1\linewidth]{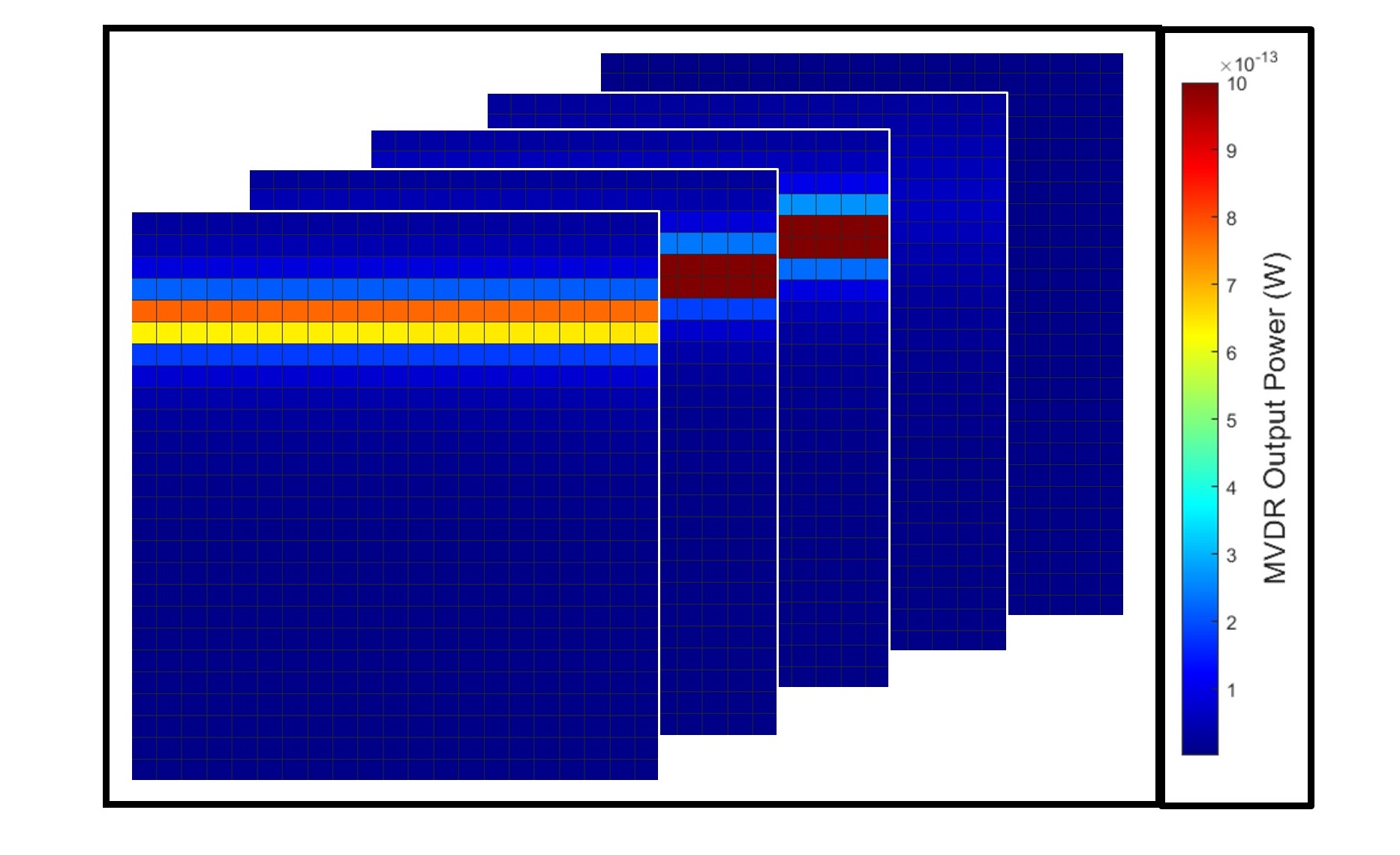}
    \caption{Representative $5\times26\times21$ heatmap tensor from the RFView example scenario (Section \ref{Sec4}). Each heatmap pertains to a unique range bin.}
    \label{heatmap}
\end{figure}

As an illustration of our approach, we generate heatmaps of the output power of an MVDR beamformer in range, azimuth and elevation angle. Consider an $L$-element receiver array for the radar receiver. Let $\mathbf{Y_r}\in \bbC^{L \times K}$ be a matrix consisting of $K$ independent realizations of the matched filtered radar array data for range bin $r$, where $r\in\mathbb{N}$. Let $\mathbf{\hat{R}_\mathbf{r}} \in \mathbb{C}^{L \times L}$ denote the sample covariance matrix obtained from $\mathbf{Y_r}$ for range bin $r$. Let $\mathbf{a_r}(\theta, \phi) \in \mathbb{C}^L$ denote the array steering vector associated with coordinates $(r, \theta, \phi)$ in range, azimuth, and elevation. This steering vector is provided by RFView. The output power, $\mathrm{P}_r(\theta,\phi) \in \mathbb{R}^+$, of the MVDR beamformer for coordinates $(r, \theta, \phi)$ is given by 
\begin{align}
\mathrm{P}_r(\theta,\phi)=\left|\frac{\mathbf{a_r}(\theta, \phi)^H\mathbf{a_r}(\theta, \phi)}{\mathbf{a_r}(\theta, \phi)^H\mathbf{\hat{R}_\mathbf{r}}^{-1}\mathbf{a_r}(\theta, \phi)} \right|.
\end{align}
Sweeping the steering vector across $\theta$ and $\phi$ at a desired resolution $(\Delta \theta,\Delta \phi)$ and recording $\mathrm{P}_r(\theta,\phi)$ at each location, we produce a heatmap image in azimuth and elevation. Stacking these images over consecutive range bins (indexed by $r$) yields a three dimensional heatmap tensor for STAP. The depth of this tensor (the number of stacked range bins) is a parameter that can be adjusted, and the exact target location information is encoded into the tensor's leakage pattern. We note that every realization of the matched filtered radar array data produces a unique heatmap tensor. Collectively, these tensors comprise the training and testing examples of our dataset. An example of this three dimensional heatmap tensor is given in Figure \ref{heatmap} for the radar simulation parameters specified in Section \ref{Sec4}.

\section{RFView Example Scenario}\label{Sec4}

\begin{figure}[hbt!]
    \centering
    \includegraphics[width=1\linewidth]{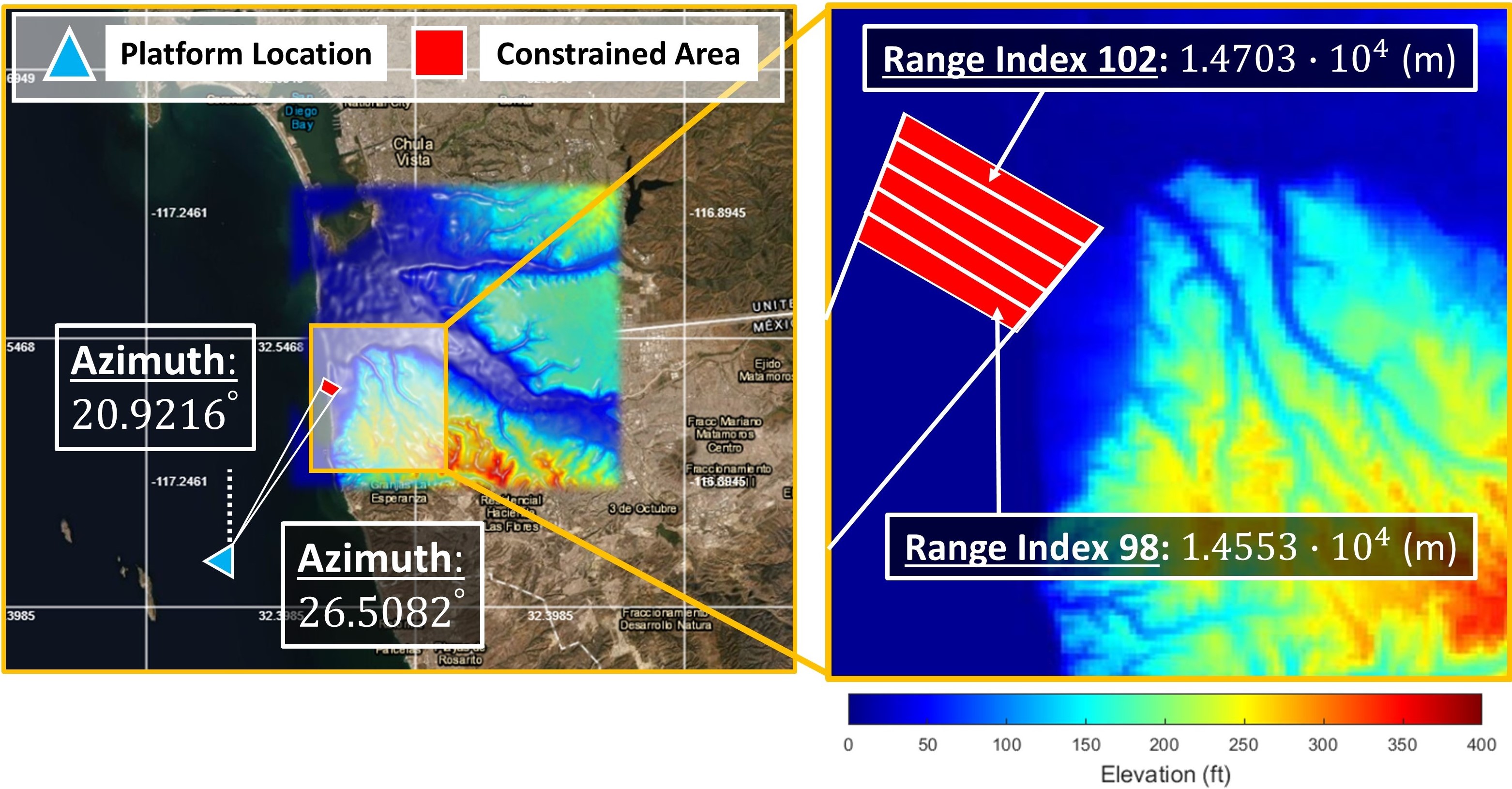}
    \caption{The map of the area of study. The blue triangle indicates the platform location and the red region is the constrained area for target placement. The elevation heatmap overlaying the left image represents the simulation region.
    \\ Left image source: USGS EarthExplorer.}
    \label{map}
\end{figure}

For our RFView example scenario, we consider an airborne radar platform flying over the coast of Southern California. The simulation region covers a $20 \ \text{km} \times 20 \ \text{km}$ area, which is uniformly divided into a $200 \times 200$ grid. Each grid cell is $100 \ \text{m} \times 100 \ \text{m}$ in size. RFView aggregates the information on land types, the geographical characteristics within each grid cell, and the radar parameters to simulate the radar return signal. The radar and site parameters from this study are given in Table \ref{radar parameters}, and the simulation region and platform location are shown in Figure \ref{map}. The radar operates in `spotlight' mode and always points toward the center of the simulation region.

\begin{table}[hbt!]
\caption{Simulation Parameters}
\label{radar parameters}
\begin{center}
\begin{tabular}{lp{4cm}}
\hline Parameter & Value \\
\hline Carrier frequency & $10,000 \ \mathrm{MHz}$ \\
Bandwidth & $5 \ \mathrm{MHz}$ \\
Pulse repetition frequency & $1100 \ \mathrm{Hz}$ \\
Duty factor & $10$ \\
Transmitter antenna & $48$ isotropic elements (horizontal) $\times \ 5$ isotropic elements (vertical)\\
Receiver antenna & $16$ horizontal elements \\
Antenna element spacing & $0.015 \ \mathrm{m}$ \\
Platform latitude & $32.4275^{\circ}$\\
Platform longitude & $-117.1993^{\circ}$ \\
Platform height & $1000 \ \mathrm{m}$ \\
Platform speed & $100 \ \mathrm{m/s}$ heading North \\
Area latitude (min,max) & $(32.4611^{\circ},32.6399^{\circ})$\\
Area longitude (min,max) & $(-117.1554^{\circ},-116.9433^{\circ})$\\
Range resolution & $30 \ \mathrm{m}$ \\
Number of range bins & $680$
\end{tabular}
\end{center}
\end{table}

We randomly place a target with an RCS that is arbitrarily selected from a uniform distribution between $75-85 \ \text{dBsm}$ within the constrained area shown in red in Figure \ref{map}. This area contains five range bins and varies in elevation angle $\phi$ from $-4.1^{\circ}$ to $-3.9^{\circ}$ and in azimuth angle $\theta$ from $20^{\circ}$ to $30^{\circ}$. The subsequent return signals have an average signal-to-clutter-to-noise ratio (SCNR) of $-2.82$ dB. We use RFView to generate $N$ radar return signals with different target locations and RCS combinations, with $N$ ranging from $10^4$ to $9 \times 10^4$ in increments of $10^4$. To isolate the range bins, we matched filter each received signal. We then evaluate the output power of an MVDR beamformer at different azimuth angle and elevation angle combinations, with resolution $(\Delta\theta=0.4^{\circ},\Delta\phi=0.01^{\circ})$ to obtain a heatmap for a given range bin $r$. Stacking images across the five range bins yields a heatmap tensor of size $5\times 26\times 21$. To calculate $\mathrm{P}_r(\theta,\phi)$,  it is necessary to compute the sample covariance, $\mathbf{\hat{R}_\mathbf{r}}$, of the received radar array data. We use $K = 100$ realizations for this procedure. The steering vector $\mathbf{a_r}(\theta, \phi)$ for each test location $(r,\theta,\phi)$ on the grid is generated from the normalized target-only response for coordinates $(r,\theta,\phi)$ from RFView. 

We save the ground truth target location for each heatmap tensor example using the standard Cartesian coordinate system with the platform at the origin, the Northward-pointing line as the x-axis, and the upward-pointing line as the z-axis. Our final dataset comprises of the heatmap tensors as the features and the true target locations as the labels which will be used by our regression CNN in Section \ref{Sec6}.

\section{Application of Computer Vision} \label{Sec5}


In the literature on computer vision, there are numerous existing classical algorithms for object detection and localization. These algorithms identify objects of interest by determining bounding boxes around familiarized entities. Spurred by the recent re-emergence of deep neural networks (DNN), many deep-learning based object detection algorithms have been proposed. Examples of these algorithms include R-CNN~\cite{girshick2014rich}, Fast R-CNN~\cite{girshick2015fast}, Faster R-CNN~\cite{ren2015faster}, You Only Look Once (YOLO)~\cite{redmon2016you}, and Single Shot MultiBox Detector (SSD)\cite{liu2016ssd} to name a few. Some of these DNN approaches, such as R-CNN, Fast R-CNN, and Faster R-CNN, typically consist of a selective search algorithm or a proposal network that suggests regions of interest (ROI) in the image/video that may contain objects of interest. Approaches such as YOLO and SSD offer an entire end-to-end solution for object detection. However, the proposed regions by the R-CNN families may not contain any object of interest. Thus a decision has to be made regarding whether an object (target) is present in a given ROI or not. This implies the existence of a classification network. Furthermore, a regression network is used to localize the bounding box around each object of interest, where every such box is selected from a set of anchor (prior) boxes.


We note that all of the aforementioned computer vision algorithms are typically trained and optimized using natural images datasets, including those containing high-resolution images such as COCO~\cite{lin2014microsoft} and Pascal VOC~\cite{pascalvoc}, and thus cannot be directly applied to radar heatmap datasets as described above. The goal of our ongoing research is to develop object detection algorithms suitable for data-driven radar target detection and localization inspired by existing computer vision algorithms. In the next section, we show the potential of our proposed approach for target localization by designing a regression convolutional neural network.

\section{Proposed Regression CNN and Empirical Results for Target Localization} \label{Sec6}

As a preliminary example, we design a regression CNN to estimate the position of a single target in the presence of clutter and noise using heatmap tensors from MVDR beamforming. We consider the dataset described in our example scenario in Section \ref{Sec4}, and we split this dataset of size $N$ such that $90\%$ of the dataset is used for training and the remaining $10\%$ is used for testing, where $N_{train} = 0.9N$ and $N_{test} = 0.1N$. We then build the regression network using a CNN architecture to learn the location of each target given its heatmap tensor.

\begin{figure}[hbt!]
    \centering
    \includegraphics[width=1\linewidth]{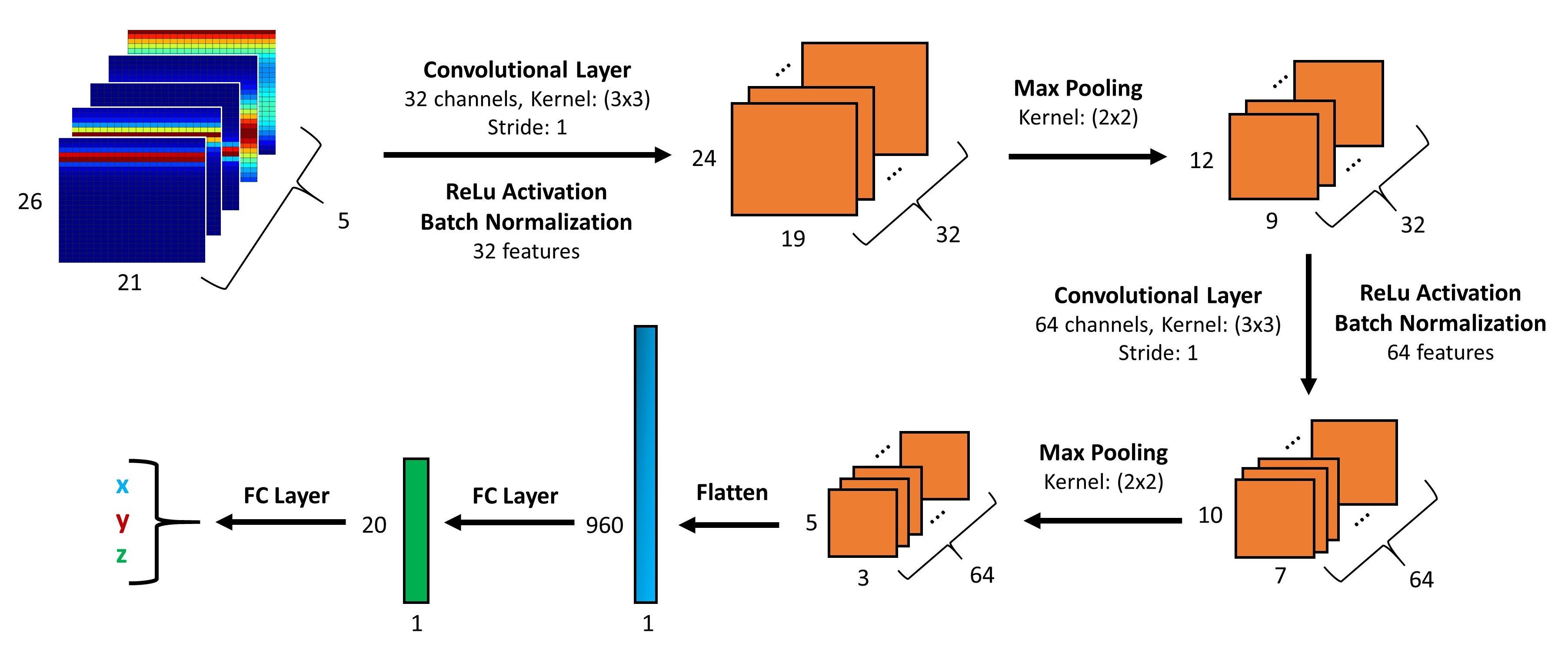}
    \caption{CNN architecture used for the regression network.}
    \label{CNN-architecture}
\end{figure}

The structure of our CNN is shown in Figure \ref{CNN-architecture}. Each of our $N_{train}$ training examples are of size $5 \times 26 \times 21$ and can be visualized as a set of five heat maps (one for each range) of size $26 \times 21$ (see Figure \ref{heatmap} for graphical representation). Our input examples pass through a 32-channel convolutional layer with kernel size $3 \times 3$ and stride 1, after which we apply the ReLu activation function and batch normalization. We apply max pooling to this output with kernel size $2 \times 2$. Then, we pass the output through another 64-channel convolutional layer with kernel size $3 \times 3$ and stride 1, once again applying ReLu activation and batch normalization. We apply max pooling to this output with kernel size $2 \times 2$, which we subsequently flatten and pass through two fully connected layers to obtain the predicted location of our target in $(x,y,z)$ coordinates.

The objective function that is minimized for this CNN is defined to be the average of the squared Euclidean distance between the predicted target locations and the ground truth locations (mean squared error), which is a popular metric for regression problems. However, to report the localization error in meters, we use the average Euclidean distance between the predicted target locations and the ground truth locations in plotting our testing errors. More specifically, we let $(x_i, y_i, z_i)$ denote the true target location for example $i$ in our data set and let $(\hat{x}_i, \hat{y}_i, \hat{z}_i)$ be the predicted location for this example from the CNN model. The testing error, $Err_{CNN}$, over $N_{test}$ test examples, is defined as follows:
\begin{align}
Err_{CNN} = \frac{\sum_{i = 1}^{N_{test}} \| (x_i, y_i, z_i) - (\hat{x}_i, \hat{y}_i, \hat{z}_i)\|_2}{N_{test}} .
\end{align}
We compare the error in target localization from our regression network with the error from a more traditional approach of using the cell with the peak MVDR output power as the predicted target location. Concretely, let $(\Tilde{x}_i, \Tilde{y}_i, \Tilde{z}_i)$ be the center of the grid cell that contains the peak output power in the MVDR heatmap for example $i$ from the dataset. Over the $N_{test}$ test examples, we can compute the error, $Err_{MVDR}$, in using this traditional method as follows:
\begin{align}
Err_{MVDR} = \frac{\sum_{i = 1}^{N_{test}} \| (x_i, y_i, z_i) - (\Tilde{x}_i, \Tilde{y}_i, \Tilde{z}_i)\|_2}{N_{test}}.
\end{align}

\begin{figure}[hbt!]
    \centering
    \includegraphics[width=1\linewidth]{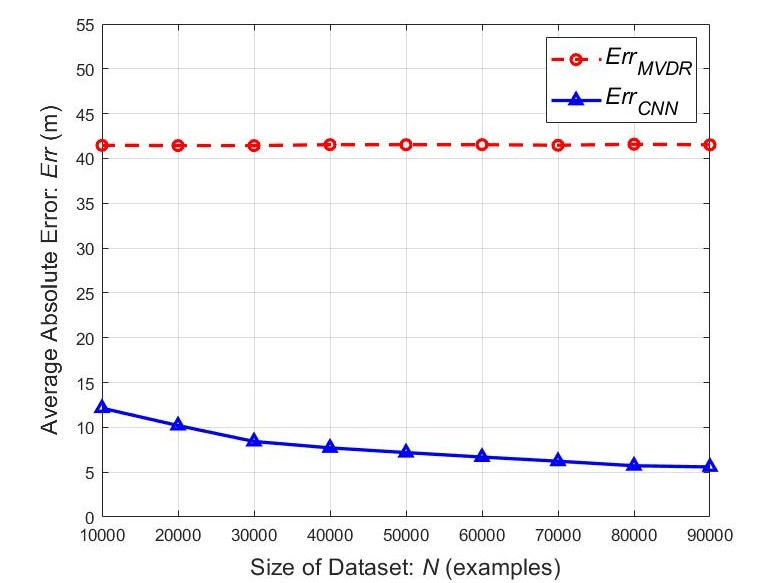}
    \caption{Comparing average absolute error for target localization using the regression CNN model (in blue) and the traditional peak-finding method with MVDR beamforming (in red), for an average SCNR of $-2.82 \ \text{dB}$.
}
    \label{avg-abs-error}
\end{figure}

Figure \ref{avg-abs-error} compares $Err_{CNN}$ and $Err_{MVDR}$ for different sizes of the dataset generated in our RFView example scenario. The figure demonstrates that our proposed regression network achieves a significantly lower error in predicting the true target location when compared to the traditional scheme of using the midpoint of the cell with the peak MVDR output power, culminating in a seven-fold improvement for $N = 9\times10^4$ examples. This regression CNN was trained using the Adam optimizer~\cite{kingma_14}. The learning rate hyperparameter, $\alpha$, was tuned by experimentation, with a final choice of $\alpha = 5 \times 10^{-4}$. 

\section{Conclusions and Future Work} \label{Sec7}

The emergence of site-specific high-fidelity radio frequency modeling and simulation tools such as RFView has made it possible to approach classical problems in radar using a data-driven methodology. In this work, we discussed our ongoing research toward data-driven STAP radar. Our approach builds upon classical STAP radar techniques and is inspired by existing computer vision algorithms for object detection. Regression networks are a key component of the aforementioned computer vision tools. To demonstrate the feasibility of our approach, we provided a regression network for target localization using heatmap tensors from MVDR beamforming and quantified its gain over the classical approach. New questions arise regarding how models trained using antenna configurations over varied elevations and topographies can be applied to unseen scenarios. Another unexplored direction concerns the received radar data: namely, the trade-off between information preservation and neural network performance. In this regard, it is necessary to strike a balance between model accuracy and computational speed.


\section*{Acknowledgment}
This work is supported in part by the Air Force Office of Scientific Research (AFOSR) under award FA9550-21-1-0235. Dr. Muralidhar Rangaswamy and Dr. Bosung Kang are supported by the AFOSR under project 20RYCORO51. Dr. Sandeep Gogineni is supported by the AFOSR under project 20RYCOR052.

\bibliographystyle{IEEEtran}
\bibliography{conference_101719}

\end{document}